# Efficient Multiscale Multimodal Bottleneck Transformer for Audio-Video Classification

Wentao Zhu

## Abstract

*In recent years, researchers combine both audio and video signals to deal with challenges where actions are not well represented or captured by visual cues. However, how to effectively leverage the two modalities is still under development. In this work, we develop a multiscale multimodal Transformer (MMT) that leverages hierarchical representation learning. Particularly, MMT is composed of a novel multiscale audio Transformer (MAT) and a multiscale video Transformer [43]. To learn a discriminative cross-modality fusion, we further design multimodal supervised contrastive objectives called audio-video contrastive loss (AVC) and intra-modal contrastive loss (IMC) that robustly align the two modalities. MMT surpasses previous state-of-the-art approaches by 7.3% and 2.1% on Kinetics-Sounds and VG-GSound in terms of the top-1 accuracy without external training data. Moreover, the proposed MAT significantly outperforms AST [28] by 22.2%, 4.4% and 4.7% on three public benchmark datasets, and is about 3× more efficient based on the number of FLOPs and 9.8× more efficient based on GPU memory usage.*

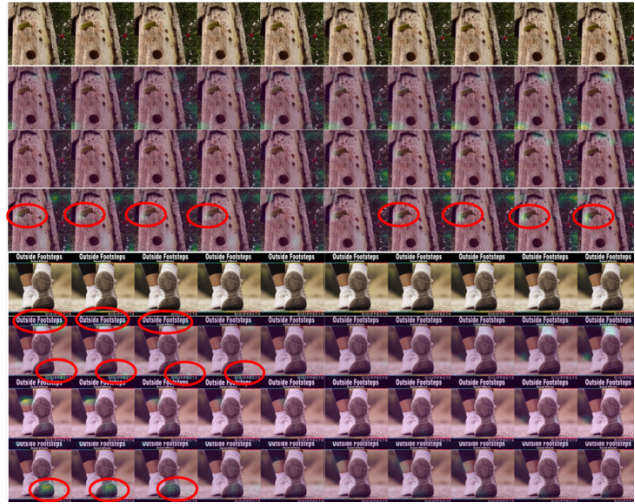

Figure 1. Two test cases, "woodpecker pecking tree" (the 1st-4th rows) and "footsteps on snow" (the 5th-8th rows), in VGGSound test set. Video only model [43] incorrectly predicts them as "playing glockenspiel" and "female singing" in the 2nd and 6th rows. AVBottleneck in § 3.4 incorrectly predicts the first case as "playing didgeridoo" in the 3rd row.

## 1. Introduction

Video-based action recognition has made tremendous progress in recent years due to the availability of massive large-scale annotated datasets and recent advances in Transformer architectures [20]. However, most existing approaches are focusing on visual cues [80, 81]. Audio signals have been leveraged together with video signals by researchers [34, 35, 79] to achieve better action recognition. For instance, Fig. 1 depicts "woodpecker pecking tree" (the 1st row) and "footsteps on snow" (the 5th row) in VG-GSound test set [13] where video only model [43] incorrectly predicts them as "playing glockenspiel" (the 2nd row) and "female singing" (the 6th row). With audio signal, our multiscale multimodal Transformer can successfully detect the sound emitting objects, *i.e.*, a woodpecker and shoes, from the large attention areas highlighted by red ovals in the 4th and 8th rows from the GradCam [58] visualizations. As can be seen, audio plays an essential role in identifying the sound emitting related actions. Visual information alone is not sufficient and can lead to misclassifications. In this work, we propose a unified and end-to-end trained multiscale architecture that effectively learns from multiple modalities for audio-video classification.

As one of the most important modalities, audio signal introduces complementary information, which can be efficiently perceived in a hierarchical structure [19, 78], *e.g.*, from each individual audio sample to audio activities and semantic audio classes. In the convolutional networks, the hierarchical feature learning can be achieved through various dilation rates and pooling strategies along the time dimension for audio classification [59]. In this work, we learn the audio features hierarchically with a multiscale Transformer structure.

In addition to audio data, the multiscale transformer architecture is also applied to video signal to extract hierarchical spatial-temporal representation. In particular, we propose a novel multimodal Transformer to extract a joint spatio-temporal and audio representation from video and audio data sources. Specifically, the multimodal Transformer efficiently conducts one-, two- and three-dimensional pooling operations along the time (and frequency), spatial and temporal dimensions in audio and video encoders. To effectively construct feature representation from multi-modality signals, contrastive learning has been widely explored in existing works [4, 72, 73]. To learn a discriminative cross-modality fusion, in this work, we propose a supervised multimodal alignment loss function, called audio-video contrastive (AVC) learning. The proposed loss aligns multimodal representations from the same category instead of the same instance in previous work. Similarly, we further incorporate the label supervision into intra-modality contrastive learning.

The **key contributions** of our work are summarized below:
- We propose a novel multiscale audio Transformer (MAT), leveraging one- and two-dimension multiscale hierarchical representation learning across the time and frequency dimensions in audio classification. MAT progressively increases the channel capacity of the intermediate latent sequence while reducing its temporal length for audio classification.
- We build a novel unified and end-to-end trained multiscale multimodal Transformer (MMT), which employs the proposed MAT and one of the current state-of-the-art video Transformers [43]. To learn compact and discriminative modality representations for multimodal feature fusion, we develop audio-video contrastive (AVC) loss and intra-modality contrastive loss considering label supervision to enhance multimodal alignment.
- Experiments on Kinetics-Sounds [6], Epic-Kitchens-100 [16] and VGGSound [13], demonstrate that MAT outperforms the previous audio Transformer [28] by 22.2%, 4.4% and 4.7% respectively, in terms of top-1 accuracy. MMT surpasses the previous state-of-the-art counterparts by 7.3% and 2.1% on Kinetics-Sounds and VGGSound without external training data.

## 2. Related Work

Learning effective audio-visual representations for video or audio classification can be improved by leveraging the natural alignment between audio and visual data [5, 9, 14, 15, 39, 49, 52–54]. Moreover, audio-visual learning has several applications such as video sound localization [1, 3, 7, 12, 25, 45, 51, 56, 60–62, 70, 74, 75], audio-visual synchronization [21], person-clustering in videos [11], (visual) speech and speaker recognition [2, 10, 48], and audio synthesis using visual information [24, 27, 37, 77].

There are several hierarchical Transformers proposed in efficient language processing and computer vision. Swin Transformer [46] designs a shifted window strategy in a hierarchical image Transformer. PVT [66] uses a progressive shrinking pyramid to reduce the computations of large feature maps for dense prediction tasks, *e.g.*, object detection and semantic segmentation. Multiscale Transformers [22, 43] adopts several channel-resolution scale stages and hierarchically expands the channel capacity while reducing the spatial resolution. Several works [28, 30, 40, 64] use Transformers for audio classification. We design a novel multiscale audio Transformer with one-dimensional and two-dimensional pooling operators along the time dimension and frequency dimension in audio spectrogram for audio classification, which achieves better accuracy than AST [28] with much more efficient number of parameters, FLOPs and GPU memory usage.

For multimodal cross-modality fusion, contrastive self-supervised learning can be used to align multimodal representation from different sources [42, 50]. Li *et al.* [42] firstly propose align before fusion using multimodal self-supervised contrastive loss to enhance vision and language representation learning. Yang *et al.* [72] introduce intra-modality contrastive learning into multimodal fusion and obtain a better accuracy. VideoCLIP [69] employs contrastive pre-training for zero-shot video-text understanding. Align and prompt [41] designs entity prompts for effective video-language pre-training. We design category discriminative cross-modality contrastive learning and intra-modality contrastive learning instead of instance discriminative contrastive learning in the multimodal Transformer.

## 3. Approach

Multiscale multimodal Transformer, as illustrated in Figure 2, has three main components: multiscale modality encoders (*i.e.*, multiscale audio Transformer and multiscale video Transformer), multi-modal fusion (*i.e.*, AVBottleneck to reduce computational complexity), and multimodal learning objectives. The multi-modal learning objectives consist of audio-video contrastive loss $L_{AVC}$, intra-modality contrastive loss $L_{IMC}$ and multimodal supervised cross-entropy loss $L_{CLS}^{AV}$.

### 3.1. Multiscale Audio Transformer

We can perceive an audio sequence in a hierarchical structure, from one signal value at each sampling time point to audio activities and an audio classification category for the whole sequence. To learn audio features hierarchically, we propose a multiscale audio spectrogram Transformer (MAT) with an audio spectrogram $X \in \mathsf{R}^{h \times T}$ as input, where $h$ is the number of triangular mel-frequency bins, and $T$ is the temporal length. We illustrate the multiscale audio spectrogram Transformer (MAT) in Figure 3. First, we obtain patch embedding from audio spectrogram with a $16 \times 16$ convolutional layer. The patch embeddings are further formulated as an embedding token matrix $A \in \mathbb{R}^{d \times N}$, where $d$ is the embedding dimension and $N$ is the number of tokens. One block of MAT can be a stack of multihead multiscale self-attention (MMSA), layer normalization (LN) and multilayer perceptron (MLP). The process can be expressed as:

$$A' = \text{MMSA}(\text{LN}(A)) + P(A),$$
$$\text{Block}(A) = \text{MLP}(\text{LN}(A')) + A', \quad (1)$$

where $P$ is a pooling operator, which can be a one-dimensional pooling along the time dimension or a two-dimensional pooling along both the time and frequency dimensions. One head in multihead multiscale self-attention (MSAttn) can be formulated as

$$Q = P_Q(AW_Q), \; K = P_K(AW_K), \; V = P_V(AW_V),$$
$$\text{MSAttn}(A) = Q + \text{Softmax}((QK^T + E^{(rel)})/\sqrt{d})V, \quad (2)$$

where $E^{(rel)}_{ij} = Q_i \cdot R_{p(i),p(j)} = Q_i \cdot (R^t_{t(i),t(j)} + R^f_{f(i),f(j)})$, $R^t$ and $R^f$ are positional embeddings along the temporal and feature axes in the spectrogram.

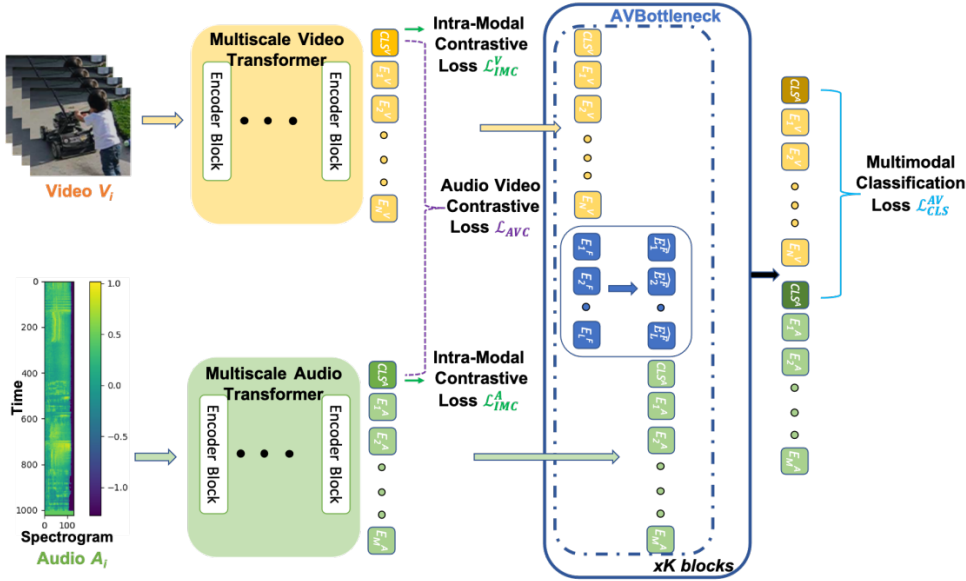

Figure 2. Illustration of multiscale multimodal Transformer, MMT, where multimodal inputs are a video clip $V_i$ and an audio spectrogram $A_i$ from the $i$-th video. Multiscale audio Transformer, MAT, learns hierarchical representations, which can effectively model the temporal dependencies. Next, we build multimodal audio-video bottleneck tokens, $\{E^F_1, \cdots, E^F_L\}$, to efficiently learn the cross-modality fusion from multiscale audio and video representations. Supervised audio-video contrastive loss $L_{AVC}$ and intra-modality contrastive loss $L_{IMC}$ encourage learning compact and discriminative representations.

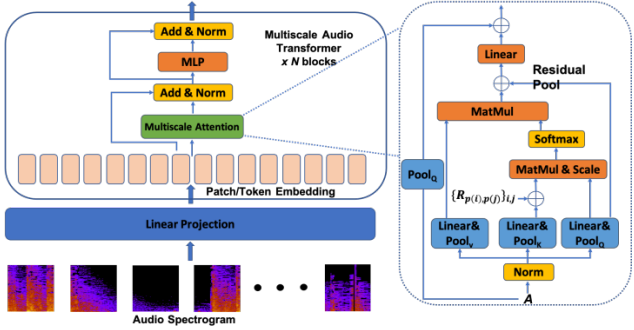

Figure 3. One block of multiscale audio Transformer (MAT). The pooling operator in the block permits to construct representations from dense to coarse resolution and is able to effectively learn hierarchical audio representations.

The multihead multiscale self-attention (MMSA) can be stacked to construct the multiscale audio spectrogram Transformer (MAT) for audio classification. The details of MAT architecture is shown in Table 1. MAT employs fewer number of feature dimensions, and fewer number of tokens than AST [28] after the first few blocks. Thus, the multiscale design leads to fewer number of parameters, FLOPs and GPU memory usage of MAT than AST. We have also explored other multiscale pooling schedules and strategies, and find that the design in Table 1 yields the best accuracy. Please refer to the supplementary for more details.

Compared with the previous audio spectrogram Transformer [28], MAT can efficiently extract representation that effectively models hierarchical characteristics of audio signals. In § 4, we demonstrate that MAT significantly reduces the required GPU memory. The efficient MAT is lightweight and can be used as a component in multimodal networks. Also, the multiscale representation learning enables a larger batch size which will be helpful for the following supervised multimodal contrastive learning.

### 3.2. Audio-Video Contrastive Learning

Multimodal inputs can naturally be considered as multiple views for the same instance in contrastive learning. Previous image-text Transformer [42] shows that the image-text contrastive loss yields a better accuracy. The cross-modality contrastive learning aligns inter-modality representations, which benefits the following cross-modality fusion. The cross-modality alignment contrastive learning can be enhanced by considering label supervision to learn compact and discriminative representations.

After multiscale audio Transformer and multiscale video Transformer, we obtain audio embeddings $\{E_{CLS}^A, E_1^A, \cdots, E_M^A\}$, and video embeddings $\{E_{CLS}^V, E_1^V, \cdots, E_N^V\}$, where $M$ is the number of audio tokens and $N$ is the number of video tokens. The audio-video contrastive loss can be formulated

$$S_{A,V} = \exp(g_A(E_{CLS}^A)^T g_V(E_{CLS}^V)),$$
$$L_{AVC} = -E_{(A,V)\in D}[y_{AV} \log \frac{S_{A,V}/\tau}{\sum_{(A,V)\in D} S_{A,V}/\tau}], \quad (3)$$

where $D$ is the multimodal input consisting of audio $A$ and video $V$ signals, $y_{AV}$ is an indicator that the current $A$ and $V$ are from the same category or not in the current batch, $\tau$ is a temperature parameter, $g_A$ and $g_V$ are linear embedding layers for audio representation $E_{CLS}^A$ and video representation $E_{CLS}^V$ respectively. The dot product $g_A(\cdot)^T g_V(\cdot)$ measures the similarity of audio and video embedding, and the supervised audio-video contrastive learning $L_{AVC}$ penalizes the distribution divergence of audio and video representations for the same category, which leads to a discriminative cross-modality representation learning.

### 3.3. Intra-Modality Contrastive Learning

The cross-modality fusion can also benefit from compact intra-modality representations. Yang *et al.* [72] employ multiple views from data augmentation to construct intra-modality contrastive loss. We further integrate *label discriminative supervision* into intra-modality contrastive loss

$$S_{V_1V_2} = \exp(g_V(E_{CLS}^{V_1})^T g_V(E_{CLS}^{V_2})),$$
$$L_{IMC}^V = -E_{(V_1,V_2)\in D}[y_{V_1V_2} \log \frac{S_{V_1V_2}/\tau}{\sum_{(V_1,V_2)\in D} S_{V_1V_2}/\tau}],$$
$$S_{A_1A_2} = \exp(g_A(E_{CLS}^{A_1})^T g_A(E_{CLS}^{A_2})),$$
$$L_{IMC}^A = -E_{(A_1,A_2)\in D}[y_{A_1A_2} \log \frac{S_{A_1A_2}/\tau}{\sum_{(A_1,A_2)\in D} S_{A_1A_2}/\tau}], \quad (4)$$

where $y_{V_1V_2}$ and $y_{A_1A_2}$ are indicators that the current $V_1$ and $V_2$, and $A_1$ and $A_2$, are from the same *category* or not in the current batch respectively. The supervised intra-modality contrastive loss enables to learn discriminative and compact modality representations.

### 3.4. Learning from Multimodal Video

**AVBottleneck** Previous cross-modality Transformers either simply concatenated multimodal representations [32], or exchanged the key and value matrices between the two modalities [29]. However, due to the huge GPU memory consumption of the existing video Transformer, we construct an audio-video bottleneck Transformer, AVBottleneck, which handles varied lengths of modality tokens efficiently as illustrated in the blue round rectangle of Figure 2. Let $\{E_1^F, \cdots, E_L^F\}$ be the initial multimodal tokens, and $L$ be the number of multimodal tokens. Without loss of generality, we omit the layer number in the denotation. One multimodal bottleneck Transformer block can be formulated as

$$E^{VF} = [E_{CLS}^V, E_1^V, \cdots, E_N^V, E_1^F, \cdots, E_L^F],$$
$$\tilde{E}^{VF} = \text{MSA}(\text{LN}(E^{VF})) + E^{VF}, \quad (5)$$
$$\hat{E}^{VF} = \text{MLP}(\text{LN}(\tilde{E}^{VF})) + \tilde{E}^{VF},$$

$$E^{FA} = [\hat{E}_1^F, \cdots, \hat{E}_L^F, E_{CLS}^A, E_1^A, \cdots, E_M^A],$$
$$\tilde{E}^{FA} = \text{MSA}(\text{LN}(E^{FA})) + E^{FA}, \quad (6)$$
$$\hat{E}^{FA} = \text{MLP}(\text{LN}(\tilde{E}^{FA})) + \tilde{E}^{FA},$$

where the initial multimodal tokens can be updated by averaging the multimodal tokens along all the AVBottleneck blocks. The audio-video bottleneck block can be stacked into $K$ blocks, and the input multimodal, video and audio tokens in the following blocks are from the previous block $E_i^F = \hat{E}_i^F$, $E_i^V = \hat{E}_i^V$, and $E_i^A = \hat{E}_i^A$ respectively.

**Computational complexity** AVBottleneck reduces the computing complexity from $O((M+N)^2)$ in merged concatenation based multimodal attention [32] to $O((M+L)^2) + O((N+L)^2) \approx O(M^2) + O(N^2)$, which is the sum of complexity in one block of audio and video Transformers approximately, since $L \ll M, N$. Here, $O(M^2)$ and $O(N^2)$ are the complexities of video and audio Transformers, where $M$ and $N$ are the numbers of tokens in the video and audio Transformers respectively.

| Block | AST [2] Feature | Arch./Param. | MAT Feature | Arch./Param. |
|---|---|---|---|---|
| Input | 1×128×1024 | 0 | 1×128×1024 | 0 |
| Patch Embed. | 768×(1212=12×101) | 768×16×16×1 | 96×(8192=32×256) | 96×7×7×1 |
| Block {0, 1} | 768×1212 | Attn-MLP | 96×8192 | Attn-MLP |
| Block 2 | 768×1212 | Attn-MLP | 192×(2048=16×128) | MMSA-MLP |
| Block {3, 4} | 768×1212 | Attn-MLP | 192×2048 | Attn-MLP |
| Block 5 | 768×1212 | Attn-MLP | 384×(512=8×64) | MMSA-MLP |
| Block {6,···,11} | 768×1212 | Attn-MLP | 384×512 | Attn-MLP |
| Block 12 | - | - | 384×512 | Attn-MLP |
| Block {13,···20} | - | - | 384×512 | Attn-MLP |
| Block 21 | - | - | 768×(256=8×32) | MMSA-MLP |
| Block {22,23} | - | - | 768×256 | Attn-MLP |

Table 1. Architecture comparison between AST [28] and MAT. MAT employs multiscale representation learning and uses 58% of the number of parameters and 35% FLOPs of AST [28] from Table 2.

Finally, we concatenate the video and audio representations $[E^V_{CLS}, E^A_{CLS}]$ and pass it through a fully connected layer for multimodal classification. The supervised multimodal loss is formulated as

$$L^{AV}_{CLS} = -\frac{1}{n} \sum_{i=1}^{n} \sum_{c=1}^{C} [y_i(c) \log p^{AV}_i(c)], \quad (7)$$

where $p^{AV}_i(c)$ is the multimodal classification probability for the $i$-th video and label index $c$. An end-to-end trained hybrid loss consisting of multimodal video classification and supervised multimodal contrastive learning objectives for the multimodal Transformer learns effectively from the training data

$$L = L^{AV}_{CLS} + \lambda_1 L_{AVC} + \lambda_2 \frac{(L^V_{IMC} + L^A_{IMC})}{2}, \quad (8)$$

where $\lambda_1$, and $\lambda_2$ are hyperparameters to balance the loss scales in the training. The inference is consistent with the training, and we use multimodal prediction $p^{AV}_i$ directly.

## 4. Experimental Results

### 4.1. Datasets

We experiment with three audio-video classification datasets – Kinetics-Sounds [6], Epic-Kitchens-100 [16–18], and VGGSound [13].

**Kinetics-Sounds** is a commonly used subset of Kinetics [33], which consists of 10-second videos sampled at 25fps from YouTube. As Kinetics-400 is a dynamic dataset and videos may be removed from YouTube, we follow the dataset collection in Xiao et al. [68], and we collect 22,914 valid training multimodal videos and 1,585 test videos.

**Epic-Kitchens-100** consists of 90,000 variable length egocentric clips spanning 100 hours capturing daily kitchen activities, which formulates each action into a verb and a noun. We employ two classification heads, one for verb classification and the other one for noun classification. The dataset mainly consists of short clips with an average length of 2.6 seconds. Following the standard protocol [16], we report top-1 action-, verb- and noun-accuracies with action accuracy being the primary metric.

**VGGSound** is a large scale action recognition dataset, which consists of about 200K 10-second clips and 309 categories ranging from human actions and sound-emitting objects to human-object interactions. Like other YouTube datasets, e.g., K400 [33], some clips are no longer available. After removing invalid clips, we collect 159,223 valid training multimodal videos and 12,790 test videos.

**Implementation details** We employ 16 frames for multiscale video Transformer and 5 ensemble views in the inference. Due to the efficiency of multiscale audio Transformer, we are able to train the multimodal model using a batch size of 64 on 8 NVIDIA A100 GPUs, each with 40 GB of memory. We set the numbers of AVBottleneck blocks $K$ and tokens $L$ as 4. $\tau$ is fixed as 0.07 and the dimensions of $g_A$ and $g_V$ are fixed as 256 following Li et al. [42]. For multiscale audio Transformer, we use audio spectrogram of size 128×1024 and ImageNet-1K pretrained weights following the same setting as AST [28]. We utilize MViTv2-B [43] as the multiscale video encoder. AdamW [47] is used in the backpropagation and the learning rate is set as 0.0001. The numbers of epochs are 50, 100, 300 on VGGSound, Epic-Kitechens-100 and Kinetics-Sounds respectively. To reduce the effort of tuning hyper-parameter, we validate the hyper-parameter one by one as shown in the ablation study while fixing previous validated hyperparameter based on validation set. We set $\lambda_1$ and $\lambda_2$ as 0.25, 0.25. These hyperparameters are generally set to balance the loss values into the same scale, and we did not tune these hyperparameters because of the long training time of each experiment. Other hyperparameters follow the recipe of MViTv2-B [43].

| Models | Modalities | | | | | |
|---|---|---|---|---|---|---|
| Chen et al. [13] | A | N/A | N/A | 48.8 | 76.5 | |
| AudioSlowFast [35] | A | N/A | N/A | 50.1 | 77.9 | |
| MBT [49] (AST) | A | 52.6 | 71.5 | 52.3 | 78.1 | |
| MAT (Ours) | A | **74.8** (22.2%↑) | **93.1** | **57.0** (4.7%↑) | **81.3** | |
| AVSlowFast, R101 [68] | A, V | 85.0 | N/A | N/A | N/A | |
| MBT [49] | V | 80.7 | 94.9 | 51.2 | 72.6 | |
| MBT [49] | A, V | 85.0 | 96.8 | 64.1 | 85.6 | |
| MMT (Ours) | A, V | **92.3** (7.3%↑) | **99.2** | **66.2** (2.1%↑) | **85.7** | |

| Models | Modalities | Verb | Noun | Action | FLOPs (G) | GPU Mem (G) |
|---|---|---|---|---|---|---|
| Damen et al. [16] | A | 42.1 | 21.5 | 14.8 | - | - |
| AudioSlowFast [35] | A | 46.5 | 22.8 | 15.4 | - | - |
| MBT [49] (AST) | A | 44.3 | 22.4 | 13.0 | 131 | 20.6×1 (bs=8) |
| MAT (Ours) | A | **50.1** | **24.2** (1.4% ↑) | **17.4** (2.0%↑) | **46.2** | **16.8×1 (bs=64)** |
| TSN [65] | V, F | 60.2 | 46.0 | 33.2 | - | |
| TRN [76] | V, F | 65.9 | 45.4 | 35.3 | - | |
| TBN [34] | A, V, F | 66.0 | 47.2 | 36.7 | - | |
| TSM [44] | V, F | 67.9 | 49.0 | 38.3 | - | |
| SlowFast [23] | V | 65.6 | 50.0 | 38.5 | - | |
| MBT [49] | V | 62.0 | 56.4 | 40.7 | 140 | |
| MBT [49] | A, V | 64.8 | 58.0 | 43.4 | 317 | |
| ViViT-L/16×2 [8] | V | 66.4 | 56.8 | 44.0 | 3410 | |
| MFormer-HR [55] | V | 67.0 | 58.5 | 44.5 | 959 | |
| MeMViT, 16×4 [67] | V | 70.6 | 58.5 | 46.2 | 59 | |
| MoViNet-A6 [38] (32 frames) | V | 72.2 | 57.3 | 47.7 | 117 | |
| MeMViT [67] (24 frames) | V | 70.6 | 58.5 | 46.2 | 89 | |
| Omnivore [26] (32 frames) | V | 69.5 | 61.7 | 49.9 | 492.8 | |
| MTV-B [71] (32 frames) | V | 67.8 | 60.5 | 46.7 | 4790 | |
| MMT (Ours) (16 frames) | A, V | 70.1 | 61.0 | 47.8 | 206 | |

Table 2. Comparison to previous related state-of-the-art on Kinetics-Sounds (upper left), VGGSound (upper right) and Epic-Kitchens-100 (bottom). We report top-1 and top-5 classification accuracy on Kinetics-Sounds and VGGSound. A: Audio, V: Visual. F: Optical flow. '-' denotes unavailability from previous work. On Epic-Kitchens-100, our method achieves the best accuracy for action recognition among methods using 16 frames.

### 4.2. Results

**Comparison to state-of-the-art** Multiscale audio Transformer (MAT) outperforms previous audio Transformer [28] by 22.2%, 4.4% and 4.7% on Kinetics-Sounds [6], Epic-Kitchens-100 [16] and VGGSound [13] in Table 2, which demonstrates that the multiscale representation learning effectively models the hierarchical characteristics in audio signals. MMT surpasses its previous state-of-the-art counterparts by 7.3% and 2.1% on Kinetics-Sounds and VGGSound based on top-1 accuracy, which shows the advantage of multiscale audio Transformer, and supervised audio-video contrastive loss and intra-modality contrastive loss. The FLOPs and #Params of our multiscale audio Transformer are 46.2G and 52M, compared with 131G FLOPs and 87M #Params in AST [28]. The multiscale audio Transformer is around 3× more efficient than AST, and the multiscale multimodal Transformer is 1.5× more efficient than previous multimodal appraoch, MBT [49], based on the number of FLOPs.

| Models | Kinetics-Sounds | | VGGSound | |
| --- | --- | --- | --- | --- |
| | Top-1 | Top-5 | Top-1 | Top-5 |
| Video Only | 91.6 | 98.8 | 56.1 | 77.9 |
| Avg | 92.0 | 99.1 | 62.4 | 84.1 |
| AVBottle w/o MAT | 91.8 | 98.8 | 56.6 | 79.2 |
| AVBottle | 91.2 | 99.1 | 63.3 | 84.1 |
| +AL | 91.4 | 99.0 | 63.5 | 84.2 |
| +AVC | 92.2 | 99.1 | 64.9 | 85.4 |
| +AVC+IM AL | 92.2 | 99.1 | 65.7 | 85.9 |
| +AVC+IMC | **92.3** | **99.2** | **66.2** | **85.7** |

Table 3. Ablation study on Kinetics-Sounds and VGGSound. AVBottle denotes AVBottleneck in § 3.4.

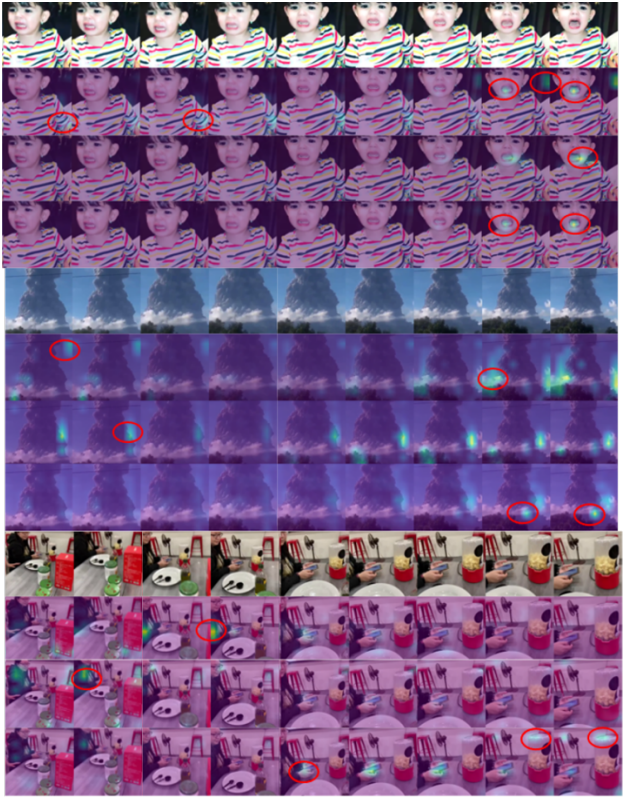

Figure 5. Visualization of three test cases in VGGSound. From top to bottom, we show 9 frames from the raw video, GradCAM [58] of video only model (MViTv2-B), AVBottleneck, MMT (ours). With well-designed strategies to learn audio and video fusion, we demonstrate that MMT can effectively understand the clip from large attention areas in the red oval highlighted circles.

**Ablation study** To verify the effectiveness of each proposed module, we progressively add each loss item to the objective. The ablations study *w.r.t.* video only (MViTv2-B [43]), simple averaging audio only and video only predictions (Avg), AVBottleneck in § 3.4, with multimodal alignment loss [42] (AL), $L_{AVC}$ (AVC), intra-modality alignment loss [72] (IM AL) and $L_{IMC}$ (IMC) are shown in Table 3 and 4 on the three datasets. From the table, we can find that 1) multimodal model outperforms one of the current state-of-the-art video Transformers [43] by a large margin, especially on VGGSound (+10.1%) and Epic-Kitchens-100 (+1.3%), 2) our multiscale multimodal Transformer with multimodal supervised contrastive learning surpasses simply fusion strategies, *i.e.*, Avg and AVBottleneck, 3) supervised multimodal contrastive losses in multimodal Transformer, *i.e.*, AVC and IMC, achieve better accuracy than their vanilla contrastive learning counterparts, *i.e.*, AL and IM AL, because the supervised contrastive learning [36] can effectively use the label supervision and learn a discriminative representation.

**Effect of multiscale audio Transformer** We compare the results of multimodal Transformer with multiscale audio Transformer and without multiscale mechanism in MAT, denoted as 'AVBottle w/o MAT', on the three datasets in Table 3 and 4. Without multiscale mechanism, the audio model takes much more GPU memory. Table 2 shows, our MAT using batch size of 64 consumes 16.8G GPU memory, which is still much less than AST [28] using batch size of 8 consuming 20.6G GPU memory on 32G V100 NVIDIA GPU. Our MAT is **9.8×** more efficient on GPU memory usage than AST. Thus, we can only use three layers, *i.e.*, the first three layers of MAT to avoid GPU out of memory issue in 8 A100 GPUs for 'AVBottle w/o MAT'. In the third layer, we maintain the same dimension of 768 as the video embedding in the bottleneck Transformer. We find that 'AVBottle w/o MAT' performs much worse than with multiscale audio Transformer on two large-scale datasets, VGGSound and Epic-Kitechens-100, which demonstrates the effectiveness of our proposed multiscale audio Transformer in the multimodal framework. Please refer to the supplementary for the ablation study on multiscale stages.

**Visualizations** We randomly pick three test clips with category names of "baby crying", "volcano explosion", and "popping popcorn" from VGGSound test set, and visualize 9 of 16 raw frames, GradCAM [58] of video only model (MViTv2-B), AVBottleneck, and the fully trained multiscale multimodal Transformer (MMT) sequentially. From

| Models | Video Only (MViTv2-B) | Avg | AVBottle w/o MAT | AVBottle | +AL | +AVC | +AVC+IM AL | +AVC+IMC |
|---|---|---|---|---|---|---|---|---|
| Verb | 67.5 | 68.7 | 69.5 | 69.8 | 69.8 | 70.0 | 69.6 | **70.1** |
| Noun | 59.2 | 59.2 | 58.5 | 59.4 | 59.9 | 60.0 | 60.1 | **61.0** |
| Action | 46.5 | 46.5 | 46.3 | 46.9 | 46.6 | 47.4 | 47.3 | **47.8** |

Table 4. Ablation study on Epic-Kitchens-100. AVBottle denotes AVBottleneck in § 3.4.

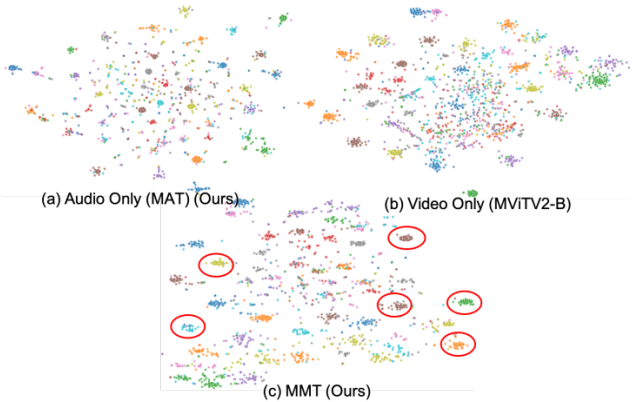

Figure 4. The t-SNE visualization [63] of representations from (a) audio only model (Our MAT), (b) video only model and (c) our MMT for random 50 categories on the test set of VGGSound. MMT learns a compact and discriminative representation, especially for red oval circled categories.

|  | Video Only | Audio Only (MAT) | MMT |
|---|---|---|---|
| ARI | 0.394 | 0.370 | **0.400** |
| HS | 0.722 | 0.718 | **0.740** |

Table 5. Clustering statistical metrics, adjusted rand index (ARI) and homogeneity score (HS), for representations on VGGSound test set. Our MMT learns a compact and discriminative representation.

the first test case (the 1-4th rows), we can find the video only model focuses on the body of the baby and incorrectly predicts this clip as "people screaming". With audio signal and supervised multimodal contrastive learning, the full MMT is able to align the audio and video well, and focuses only on the mouth of the baby to obtain the correct prediction. From the second test case (the 5-8th rows), we find that AVBottleneck in the 7th row cannot capture the fog and mountain, and it incorrectly predicts the clip as "mouse clicking". From the third case (the 9-12th rows), we find that the video only model does not have any attention on the pop-corn machine and only pays attention to human and the background table, and incorrectly predicts the clip as

"eating with cutlery", whereas MMT with audio signal and discriminative loss can fully interpret the underlying action.

We also employ t-SNE [63] to visualize the feature representations from the second to the last layer in multiscale audio Transformer (a), multiscale video Transformer (b), and our multiscale multimodal Transformer (c) on VGGSound dataset in Figure 4. For clarity, we randomly choose 2,000 test samples and 50 categories in the visualization. From the figure, we can find that our MMT learns a compact and discriminative representation, especially for red oval circled categories. In Table 5, we compare the feature representations for all the categories using two statistical metrics on VGGSound test set. The adjusted rand index (ARI) [31] computes a similarity measure between the clusters and the ground truth categories. The homogeneity score (HS) [57] checks if a cluster contains samples belonging to a single class. Both metrics can be used to evaluate the compactness and correctness of representation learning methods, and a higher value means a better model. From the table, MMT achieves the best score based on the two metrics, which validates that MMT with supervised contrastive learning can effectively learn from audio and video data sources.

## 5. Conclusion

In this work, we have proposed an effective and efficient multiscale audio Transformer, MAT, for audio classification, as well as a multiscale multimodal Transformer, MMT, for multimodal action recognition. MMT leverages advanced multiscale Transformers, *supervised* audio-video contrastive loss and intra-modal contrastive objective to efficiently learn a discriminative multimodal representation. Experimental results demonstrate that, MAT is about ⅓× more efficient based on the number of FLOPs and 9.8× more efficient on GPU memory usage, and is able to outperform AST [28] by 22.2%, 4.4% and 4.7% based on top-1 accuracy on Kinetics-Sounds, Epic-Kitchens-100 and VGGSound. MMT surpasses its previous state-of-the-art counterparts by 7.3% and 2.1% on Kinetics-Sounds and VGGSound without external training data.

*ceedings of the European Conference on Computer Vision (ECCV)*, pages 803–818, 2018. 6

[77] Hang Zhou, Ziwei Liu, Xudong Xu, Ping Luo, and Xiaogang Wang. Vision-infused deep audio inpainting. In *Proceedings of the IEEE International Conference on Computer Vision*, 2019. 2

[78] Wentao Zhu, Mohamed Omar. Multiscale Audio Spectrogram Transformer for Efficient Audio Classification. In ICASSP, 2023.

[79] Wentao Zhu. Efficient Selective Audio Masked Multimodal Bottleneck Transformer for Audio-Video Classification. arXiv, 2024.

[80] Wentao Zhu, Cuiling Lan, Junliang Xing, Wenjun Zeng, Yanghao Li, Li Shen, Xiaohui Xie. Co-occurrence feature learning for skeleton based action recognition using regularized deep LSTM networks. AAAI, 2016.

[81] Xuefan Zha, Wentao Zhu, Tingxun Lv, Sen Yang, and Ji Liu. Shifted chunk transformer for spatio-temporal representational learning. In *Advances in Neural Information Processing Systems, 2021.*
12